\documentclass[10pt]{article}
\usepackage[preprint]{tmlr}
\usepackage{amsmath,amssymb}
\usepackage{graphicx}
\usepackage{tikz}
\usetikzlibrary{arrows.meta,positioning}
\usepackage{booktabs}
\usepackage{hyperref}
\usepackage{url}
\hypersetup{pdftitle={Supervised Hebbian Class Networks: Learning Without Backpropagation},pdfauthor={John Loverich}}

\newcommand{\argmax}{\operatorname*{arg\,max}}
\title{Supervised Hebbian Class Networks:\\Learning Without Backpropagation}
\author{John Loverich\\\href{mailto:john.loverich@gmail.com}{\nolinkurl{john.loverich@gmail.com}}\\Code: \url{https://github.com/jloveric/hebbian-networks}}
\begin{document}
\maketitle

\begin{abstract}
This paper introduces Supervised Hebbian Class Networks as a method for training multilayer perceptrons and convolutional networks without gradient calculations or backpropagation. Each neuron in a learned layer is assigned a class, while connections follow the usual fully connected or convolutional pattern. The method combines a class-based supervision rule shared across layers with a Lion optimizer adapted to Hebbian activity products. All layers train together throughout the run, and their activities provide class predictions directly, without a separately trained classifier. The experiments test the method on MNIST with a multilayer perceptron and on CIFAR-10 with a convolutional network.
\end{abstract}

\section{Introduction}
Supervised Hebbian Class Networks assign each neuron in every learned layer to a class. That assignment determines how the label guides learning at the neuron's incoming connections, while the connections themselves follow the usual fully connected or convolutional pattern.

For example, in a digit classifier, some neurons are assigned to digit 0, some to digit 1, and so on. An MLP keeps its usual connections: every neuron receives the preceding layer's outputs, regardless of their assigned classes. A convolutional network keeps its usual spatial connections and shared filters; all neurons using the same filter are assigned the same class.

Prediction uses the network's activities directly. In each layer, competition keeps a prescribed number of the largest activities in absolute value among neurons assigned to each class and sets the others to zero. The network adds their mean squared activities across layers and predicts the class with the largest score. For a correct prediction, the learning rule skips that image if its class score is sufficiently above the next-highest score. Otherwise, the rule reinforces connections entering neurons assigned to the correct class and suppresses those entering neurons assigned to the highest-scoring incorrect class. Every connection's update uses the activities at its two ends.

All layers train jointly. Each minibatch passes through the complete network before any weights change. The learning rule is then applied in every layer using the recorded activities. All layers train together for the entire run, rather than being trained and frozen one at a time.

The paper contributes a supervised learning rule and an adaptation of the Lion optimizer for Hebbian learning. The supervision rule uses the relative gap between the correct-class score and the strongest incorrect-class score to decide whether to learn, then applies positive and negative learning signs to those classes throughout the network. The adapted optimizer applies Lion's running-average and sign operations to local activity products and uses the receiving neuron's activity to control weight growth. Together with competition and direct class scores, these components train all learned layers jointly without derivatives or backward error signals. Class assignments and joint Hebbian training have precedents, as discussed below; the contribution is the specific supervision and weight-update construction. The experiments demonstrate it with fully connected layers on MNIST and convolutional layers on CIFAR-10.

\subsection{Related work}
Classical Hebbian learning changes a weight in proportion to the product of its incoming and outgoing activities. Oja's rule adds an activity-dependent term to control weight growth \citep{oja1982}, and competition can determine which hidden neurons learn from an input \citep{krotov2019}. Supervised Hebbian Class Networks use both positive and negative activity products, together with competition and Lion adapted to Hebbian learning.

Multilayer SoftHebb learns features through unsupervised Hebbian competition \citep{journe2023}. Its reported accuracies use successive layer training followed by a separately trained, gradient-based linear classifier; its appendix also reports similar results when all feature layers learn simultaneously. The method presented here adds class assignments and supervision throughout the learned layers, with classification directly from their activities.

Class assignments in hidden layers are not new to Hebbian learning. \citet{amato2019} assign intermediate convolutional filters to classes, alongside filters shared by all classes. Their supervised rule substitutes label-defined teacher values for the receiving neuron's activity. Supervised Hebbian Class Networks retain the measured activities and use the label and class scores to choose whether their products reinforce or suppress a connection. Supervised Hebbian Continual Representation Learning also assigns groups of neurons to classes and predicts by summing their activities \citep{morawiecki2023}. It uses a single learned layer, freezing existing neurons and adding new ones as classes arrive.

In spiking networks, \citet{lemouel2014} use class-specific neuron pools and a margin that stops learning once the pools meet their response thresholds. Their Hebbian and anti-Hebbian updates are derived from approximate gradient descent on a margin-based loss; the thresholds constrain each pool's response rather than the gap between the correct and strongest incorrect class scores used here. Selecting the strongest incorrect class also has precedents in multiclass margin methods, including passive-aggressive learning \citep{crammer2006}.

Paired Competing Neurons assign two competing neurons to each class, encouraging specialization for target and non-target examples \citep{goupy2024pcn}. Neuronal Competition Groups use several neurons per class with competition within each group \citep{goupy2024}. Both use supervised spike-timing-dependent learning in a classification layer following an unsupervised feature extractor. CoLaNET organizes neurons into class-specific columns and combines anti-Hebbian and dopamine-modulated learning, but only its lowest layer has plastic input connections \citep{kiselev2024}.

Deep Counterstream Associative Networks also perform supervised classification using local Hebbian-type learning \citep{knoblauch2026}. Following an error, correcting target activities travel backward through learned connections to meet forward activities in a hidden layer. Supervised Hebbian Class Networks obtain all learning observations from one forward calculation and share class-dependent learning signs without propagating target activities backward.

Supervised Spike Agreement-Dependent Plasticity also learns without gradients, using agreement between a hidden neuron's spikes and the correct-class output neuron's spikes to guide hidden-layer learning \citep{sadp2026}. It demonstrates one hidden layer whose neurons have no fixed class assignments. Its teaching signal depends on output-layer activity, whereas the rule here uses activities at each connection and shared class-dependent learning signs.

The Forward-Forward algorithm uses positive and negative examples and layer activity scores, with local gradient calculations in place of end-to-end backpropagation \citep{hinton2022}. Hebbian Forward-Forward (HebbFF) replaces those gradients with activity products and also updates all layers within each epoch \citep{krutsylo2025}. It appends labels at each layer, samples incorrect labels, and weights updates using each layer's own score. Supervised Hebbian Class Networks need no candidate-label inputs: one forward calculation supplies all class scores, and their combined values choose the same correct and strongest incorrect groups for every layer.

Class groups and connections that mix their features also appear in channel-wise competitive Forward-Forward learning \citep{papachristodoulou2024}. That method trains using layer-specific losses and an interleaved schedule. A later extension, Channel-wise Spatial Gradient learning, retains class-assigned channels and interprets its updates using a Hebbian factor, but computes local loss gradients and trains layers in succession \citep{dupuis2026}. Self-Contrastive Forward-Forward (SCFF) instead learns features from image pairs without labels, trains layers in succession, and uses a separately trained linear classifier \citep{chen2025scff}.

The adapted update rule borrows the running-average and sign operations of Lion \citep{chen2023}, applying them to Hebbian activity products rather than gradients.

\section{Supervised Hebbian Class Networks}
\label{sec:model}
For a task with $C$ classes, every neuron in each of the $L$ learned layers has a fixed class assignment. A \emph{class group} is the set of neurons assigned to the same class within a layer. In an MLP, layer $\ell$ has $n_\ell$ neurons per class and $Cn_\ell$ neurons in total. In a convolutional network, it has $n_\ell$ filters per class and $Cn_\ell$ output channels in total; all spatial copies of a filter have the same class assignment. Training changes the connection weights while keeping these assignments fixed.

\begin{figure}[t]
\centering
\begin{tikzpicture}[font=\small,>=Stealth,
 neuron/.style={draw,circle,minimum size=.65cm,inner sep=1pt},
 input/.style={draw,rectangle,fill=gray!8,minimum size=.65cm,inner sep=1pt}]
\node[font=\bfseries] at (0,2.6) {Input};
\node[font=\footnotesize] at (0,2.2) {No class assignment};
\node[font=\bfseries] at (3.2,2.6) {Learned layer 1};
\node[font=\bfseries] at (6.4,2.6) {Learned layer 2};
\foreach \i/\y/\class/\shade in {1/1.5/1/blue!12,2/.6/1/blue!12,3/-.3/2/orange!18,4/-1.2/2/orange!18} {
 \node[input] (x\i) at (0,\y) {$x_{\i}$};
 \node[neuron,fill=\shade] (a\i) at (3.2,\y) {$\class$};
 \node[neuron,fill=\shade] (b\i) at (6.4,\y) {$\class$};
}
\foreach \i in {1,2,3,4} {
 \foreach \j in {1,2,3,4} {
  \draw[->,gray!65] (x\i.east) -- (a\j.west);
  \draw[->,gray!65] (a\i.east) -- (b\j.west);
 }
}
\node[align=left,anchor=west,text width=3.7cm] at (7.4,.2) {Squares hold input values $x_i$, such as image pixels.\\[5pt]Each circle is a learned neuron; its number is its assigned class.\\[5pt]Every neuron receives all values from the preceding layer.};
\end{tikzpicture}
\caption{An MLP with unassigned input coordinates and class-assigned neurons in its learned layers. Four input coordinates, two classes, and two learned layers are shown for illustration. Class assignments determine competition, scoring, and learning signs; connections follow the usual fully connected pattern. In the convolutional version, input pixels likewise have no class assignment, while each learned filter has an assigned class shared by all neurons using it.}
\label{fig:mixed}
\end{figure}
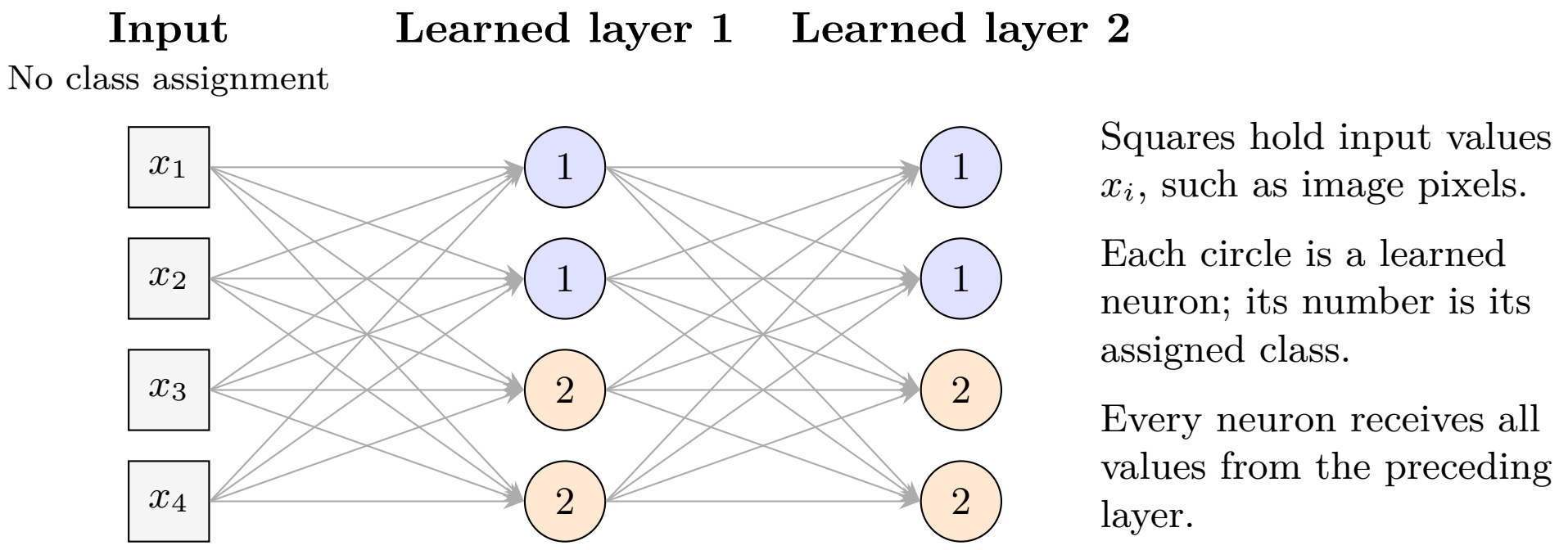

Figure~\ref{fig:mixed} shows an MLP with classes assigned to the neurons in its learned layers. The input coordinates carry image values and have no class assignments; every neuron in the first learned layer receives all input coordinates. The learned layers are fully connected as usual. In the convolutional version, each filter spans a spatial patch and all input channels. A neuron's assigned class is part of the model; the correct label of an image is used only during training and is never an input to prediction.

\subsection{Activities and competition}
For neuron $j$ in group $k$, write its weighted input as
\begin{equation}
 z_j^k=\sum_i w_{ij}^k u_i.
 \label{eq:weighted-sum}
\end{equation}
Here $u_i$ is an incoming activity and $w_{ij}^k$ is its connection weight. The index $i$ ranges over the preceding layer's neurons in an MLP, or the input patch's positions and channels in a convolution. A constant input represents the learned bias.

\paragraph{Output positions in a convolution.}
A convolution applies the same filter at many positions in an image. For example, sliding a $3\times3$ filter over a $32\times32$ image, one pixel at a time and without padding, produces a $30\times30$ output. Each of these 900 output positions corresponds to one placement of the input patch. The index $p$ identifies that placement, not a pixel within the patch. A single index is sufficient to enumerate the output grid: $p=1$ can denote the upper-left position, $p=2$ the next position to its right, and so on across each row.

The three indices distinguish different parts of this calculation:
\begin{itemize}
 \item $p$: which patch placement, or output position, is being processed.
 \item $i$: which input coordinate within that patch is used, including its channel. A $3\times3$ RGB patch contains 27 pixel values; the additional constant coordinate represents the bias.
 \item $j$: which filter within class group $k$ is applied.
\end{itemize}
For example, $i$ might identify the red value at the upper-left corner of a patch. As the patch moves to a new position $p$, that input value changes, but its connection weight in filter $j$ remains the same. Each filter therefore produces one activity at each output position. In deeper layers, the patches contain activities from the preceding layer rather than image pixels.

Within each group, competition keeps the $K_\ell$ outputs with the largest absolute values and sets the others to zero. The resulting output is $h_j^k$:
\begin{equation}
 h_j^k=\begin{cases}
 z_j^k,&j\text{ is among the selected neurons or channels},\\
 0,&\text{otherwise}.
 \end{cases}
 \label{eq:forward}
\end{equation}
Competition preserves the signs of the outputs it keeps. For example, $(-3,0.2,2,-0.1)$ with $K_\ell=2$ becomes $(-3,0,2,0)$. This selection supplies the network's nonlinearity; no additional activation function is applied. In a convolutional layer, competition selects $K_\ell$ channels within each class group separately at every output position $p$. It selects which filters respond at a given position, not which positions in the image to retain. Prediction and training use the same selection rule.

\subsection{Class scores}
For image $x$, the notation $h_{\ell,p,j}^k(x)$ makes the layer $\ell$ and output position $p$ explicit; $j$ and $k$ still identify the neuron or filter and its class group. Let $P_\ell$ be the number of output positions in that layer. In the example above, $P_\ell=30\times30=900$. An MLP has no sliding patches: $j$ identifies a neuron within its class group, and $p$ takes just one value, so $P_\ell=1$.

Each class score adds the layer contributions. Within a layer, it averages $h^2$ across the $K_\ell$ selected neurons or channels and the $P_\ell$ output positions:
\begin{equation}
 S_k(x)=\sum_{\ell=1}^{L}\frac{1}{P_\ell K_\ell}
              \sum_{p=1}^{P_\ell}\sum_{j=1}^{n_\ell}
              \bigl(h_{\ell,p,j}^k(x)\bigr)^2,
 \qquad \widehat y(x)=\argmax_k S_k(x).
 \label{eq:score}
\end{equation}
An activity of $-2$ contributes four to the score, just as $+2$ does, although their signs remain different in the values passed forward and used for learning. The denominator $P_\ell K_\ell$ is the number of output positions multiplied by the number of neurons or channels selected per class at each position. These class scores measure neural activity; they are not probabilities. The score has no learned coefficients and needs no separate classifier.

\section{Learning from local activity}
\label{sec:learning}
\begin{figure}[!ht]
\centering
\begin{tikzpicture}[font=\small,>=Stealth,
 input/.style={draw,rectangle,minimum width=1.25cm,minimum height=.65cm,fill=white},
 neuron/.style={draw,circle,minimum size=.75cm,inner sep=0pt,fill=white},
 output/.style={draw,rounded corners=2pt,minimum width=1.65cm,minimum height=.65cm,fill=white}]
\draw[rounded corners=5pt,draw=blue!45!black,fill=blue!4] (3.25,3.45) rectangle (12.55,6.95);
\draw[rounded corners=5pt,draw=orange!60!black,fill=orange!5] (3.25,-.15) rectangle (12.55,3.35);
\node[anchor=west,font=\bfseries] at (3.5,6.65) {Class 1 group};
\node[anchor=west,font=\bfseries] at (3.5,3.05) {Class 2 group};
\node[font=\bfseries] at (0,7.55) {Inputs};
\node[font=\bfseries] at (4,7.55) {Neurons};
\node[font=\bfseries] at (8,7.55) {Competition};
\node[font=\bfseries] at (11.45,7.55) {Outputs};
\foreach \i/\y/\label in {1/5.7/{u_1},2/4.2/{u_2=2},3/2.7/{u_3},4/1.2/{u_4}}
  \node[input] (in\i) at (0,\y) {$\label$};
\foreach \k/\j/\y in {1/1/6,1/2/5,1/3/4,2/1/2.4,2/2/1.4,2/3/.4}
  \node[neuron] (n\k\j) at (4,\y) {$\sum$};
\foreach \i in {1,2,3,4}
  \foreach \k in {1,2}
    \foreach \j in {1,2,3}
      \draw[->,black!25,thin] (in\i.east) -- (n\k\j.west);
\draw[->,very thick,blue!70!black] (in2.east) -- node[pos=.55,above,sloped,fill=white,inner sep=2pt] {$w_{21}^{1}$} (n11.west);
\foreach \y in {5,1.4} {
  \draw[rounded corners=3pt,fill=white] (6.8,\y-1.35) rectangle (9.2,\y+1.35);
  \node[align=center,text width=2.1cm] at (8,\y) {Keep the two\\largest $|z|$\\[5pt]Set the other\\output to zero};
}
\foreach \k/\j/\y/\z/\h in {1/1/6/3/3,1/2/5/1/0,1/3/4/-2/-2,2/1/2.4/.5/0,2/2/1.4/-4/-4,2/3/.4/2.5/2.5} {
  \draw[->] (n\k\j.east) -- node[above] {$z_{\j}^{\k}=\z$} (6.8,\y);
  \node[output] (out\k\j) at (11.45,\y) {$h_{\j}^{\k}=\h$};
  \draw[->] (9.2,\y) -- (out\k\j.west);
}
\node[align=center] at (6.1,-.75) {Each neuron computes $z_j^k=\sum_i w_{ij}^k u_i$. Competition acts separately within each class group.};
\end{tikzpicture}
\caption{One layer with four inputs and two class groups, each containing three neurons. Every input connects to every neuron. Each $\sum$ denotes a neuron's weighted sum; the example values beside the arrows are its activities before competition. With $K=2$, competition keeps the two largest absolute activities in each group, preserving their signs, and sets the third to zero. The resulting outputs $h_j^k$ pass to the next layer and are also used for learning. The highlighted connection uses input $u_2=2$ and output $h_1^1=3$ in its Hebbian product.}
\label{fig:hebbian-product}
\end{figure}
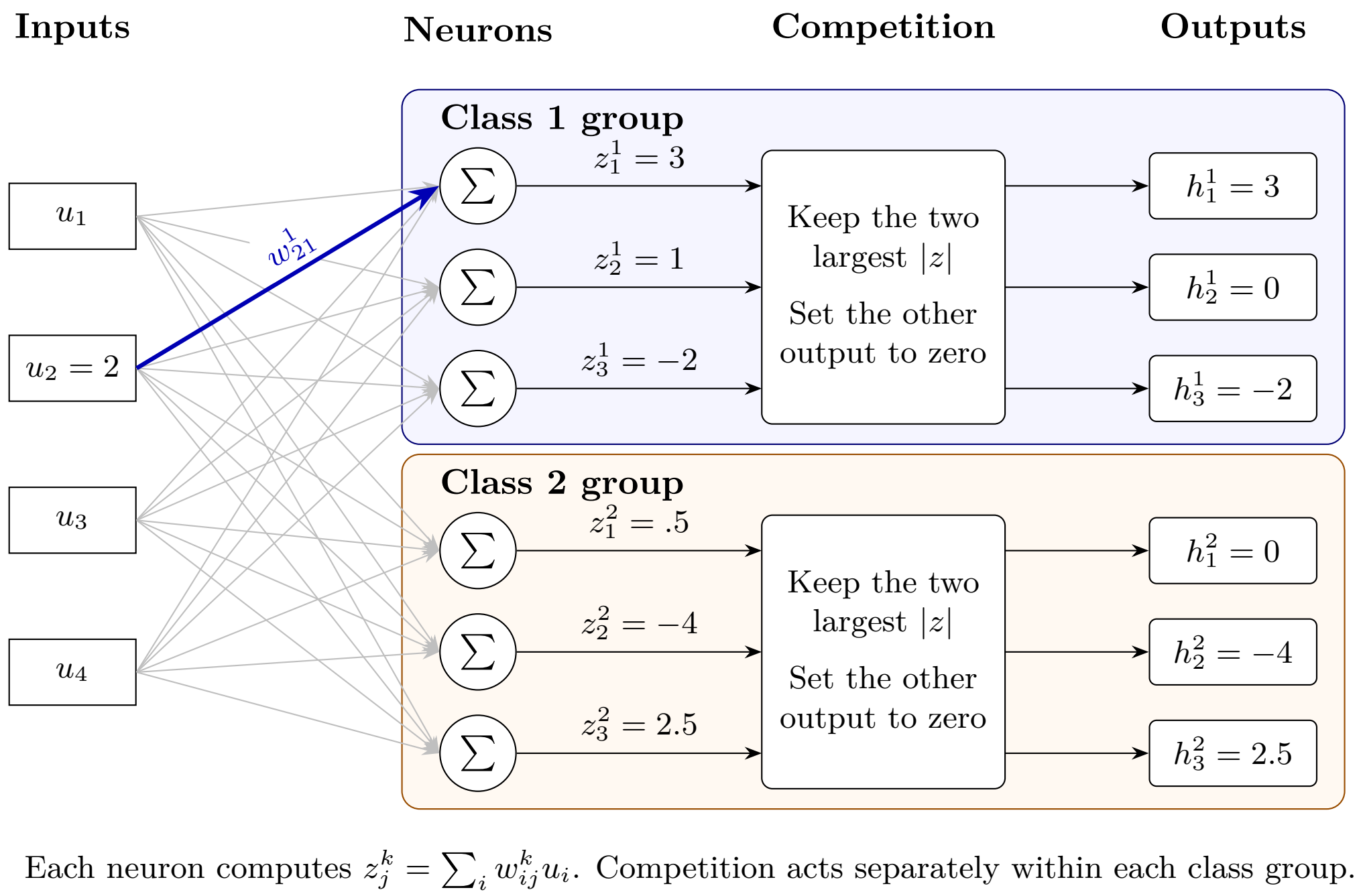

Consider one connection with weight $w_{ij}$, as shown in Figure~\ref{fig:hebbian-product}. During the forward calculation, input $u_i$ passes through this connection to neuron $j$. The neuron combines all its weighted inputs, and competition determines its output $h_j$. Learning then multiplies two values from that calculation: the input $u_i$ and the output $h_j$. This product, $u_i h_j$, is the Hebbian observation for that connection.

For the highlighted connection in Figure~\ref{fig:hebbian-product}, $u_2=2$ and $h_1^1=3$, so the product is six. Basic Hebbian learning would add $6\eta$ to $w_{21}^1$, where $\eta$ is a positive step size. The supervised rule described below first assigns a positive or negative sign to the product, averages the observations, and controls the size of the resulting weight change.

\subsection{Choosing the learning signs}
For a correctly classified image, the rule skips learning if the correct class's score is sufficiently higher than the next-highest score. Incorrect predictions, and correct predictions with too small a gap, contribute to weight updates. To specify this rule, let $y$ be the image's correct label and $r$ its highest-scoring incorrect class. Define the relative score gap, or margin, as
\begin{equation}
 \gamma=\frac{S_{y}(x)-S_{r}(x)}
                   {S_{y}(x)+S_{r}(x)+\varepsilon},
 \qquad \varepsilon=10^{-30}.
 \label{eq:margin}
\end{equation}
If the prediction is correct and $\gamma\ge\delta$, the image makes no contribution to weight updates. Here $\delta>0$ is the required margin. Otherwise, neurons assigned to the correct class use sign $s_y=+1$ and neurons assigned to the highest-scoring incorrect class use sign $s_r=-1$. All other signs are zero. These signs determine whether each receiving neuron's local activity products are added or subtracted.

The signs apply to the \emph{receiving} groups in every layer. A connection from a dog-associated neuron to a cat-associated neuron therefore uses the cat group's sign. Its update still depends on the actual activities at both ends of that connection. Positive and negative observations have equal weight.

\subsection{Observations from one image}
\label{sec:single-image}
The learning rule is first described for one image at a time. Its minibatch form is described separately in Section~\ref{sec:minibatching}; that section is not needed to follow the single-image rule below.

Fix one layer and a receiving class group $k$. Their indices are omitted from the weights and activities to keep the notation simple. Two quantities are used to update each weight $w_{ij}$: the signed activity product $A_{ij}$ and the receiving neuron's squared activity $Q_j$.

\paragraph{Fully connected layer.}
For a neuron $j$ whose output $h_j$ is nonzero and whose class has learning sign $s_k\ne0$, these quantities are simply
\begin{equation}
 A_{ij}=s_k u_i h_j, \qquad Q_j=h_j^2.
 \label{eq:single-image}
\end{equation}
For example, if $u_i=2$, $h_j=3$, and $s_k=+1$, then $A_{ij}=6$ and $Q_j=9$. With negative supervision, $A_{ij}=-6$ while $Q_j$ remains nine. The first quantity supplies the direction of learning; the second controls weight growth, as described in Section~\ref{sec:weight-update}. If $h_j=0$ or $s_k=0$, the neuron keeps its weights and stored history unchanged.

\paragraph{Convolutional layer: averaging within one image.}
A filter uses the same weights at many positions in an image. For each filter, the rule averages over positions where its output after competition is nonzero. Let $\mathcal P_j=\{p:h_{pj}\ne0\}$ be this set, and let $u_{pi}$ be coordinate $i$ of the patch at position $p$. For a filter with $s_k\ne0$ and nonempty $\mathcal P_j$, the averages are
\begin{equation}
 A_{ij}=\frac{1}{|\mathcal P_j|}\sum_{p\in\mathcal P_j}s_k u_{pi}h_{pj},
 \qquad
 Q_j=\frac{1}{|\mathcal P_j|}\sum_{p\in\mathcal P_j}h_{pj}^2.
 \label{eq:spatial}
\end{equation}
Zero outputs contribute nothing to either sum. The choice matters in the denominator: if a filter has nonzero outputs at 12 of 100 positions, both sums are divided by 12, not 100. An exactly zero output is excluded from this count even if the filter was selected by competition at that position. Thus the rule averages over nonzero responses, rather than over all positions where the filter was applied.

An alternative would be to average over positions where competition selected the filter, counting selected outputs even when they are exactly zero. The two choices give the same averages when all selected outputs are nonzero; otherwise, their denominators differ. The reported experiments use the nonzero-output count in Equation~\ref{eq:spatial}.

The input coordinate $i$ stays fixed while $p$ ranges over patch placements; different weights within the filter are not averaged together. This averaging is part of the convolutional learning rule even when training on one image at a time. If no position has a nonzero output, or if $s_k=0$, the filter's weights and history stay unchanged.

In both architectures, $A_{ij}$ can be positive or negative, while $Q_j$ is always nonnegative. Negative supervision reverses the sign of the activity product, but leaves $Q_j$ unchanged in both updates. For the reflected filters described in Section~\ref{sec:experiments}, inputs in Equation~\ref{eq:spatial} are placed in the coordinates of the selected filter orientation before their products are accumulated.

\subsection{Lion adapted to Hebbian learning}
\label{sec:weight-update}
Each connection stores a running average $m_{ij}$ of its signed products. Combine this history with the current observation to obtain a direction:
\begin{equation}
 D_{ij}=\frac{\operatorname{sign}(0.125m_{ij}+0.875A_{ij})}
                   {\sqrt{d_{\rm in}}},
 \qquad
 m_{ij}^{+}=0.99m_{ij}+0.01A_{ij}.
 \label{eq:direction}
\end{equation}
The superscript $+$ denotes the new value. The direction uses the old history, and $d_{\rm in}$ counts incoming coordinates, including the bias. The sign function returns $-1$, zero, or $+1$. This adapts Lion's sign-and-history construction \citep{chen2023}: the current Hebbian observation and its running average take the places of the gradient and gradient history. Separate averaging rates determine the update direction and the stored history. No gradient is calculated.

The experiments use two forms of the weight update. In runs using semi-implicit Lion, the update is
\begin{equation}
 w_{ij}^{+}=\frac{w_{ij}+\eta D_{ij}}{1+\eta Q_j}
 \qquad\text{(semi-implicit)}.
 \label{eq:update}
\end{equation}
In runs using explicit Lion, the update is
\begin{equation}
 w_{ij}^{+}=w_{ij}+\eta D_{ij}-\eta Q_jw_{ij}
 \qquad\text{(explicit)}.
 \label{eq:update-explicit}
\end{equation}
Both forms use the same Lion direction $D_{ij}$ and squared-activity observation $Q_j$. In Equation~\ref{eq:update}, the numerator moves the weight in the chosen direction and the denominator controls weight growth for both positive and negative observations. Unlike Lion's usual weight decay, this control depends on the receiving neuron's observed activity.

The difference is which weight appears in the activity-dependent term. The explicit form uses the old weight $w_{ij}$. The semi-implicit form uses the new weight: writing
$w_{ij}^{+}=w_{ij}+\eta D_{ij}-\eta Q_jw_{ij}^{+}$
and collecting the terms in $w_{ij}^{+}$ gives Equation~\ref{eq:update}. The Lion direction and measured activities remain fixed in both forms. The semi-implicit denominator is always positive and at least one. Dividing by this denominator cannot reverse the sign of the numerator; it does not prove convergence of the full learning process. Both forms use the same step-size schedule.

\subsection{Joint training and the scope of locality}
\noindent\begin{minipage}{\linewidth}
For one training image, the complete learning procedure has four steps:
\begin{enumerate}
 \item Pass the image through every layer using the current weights. Save the inputs $u$ and outputs $h$ at each layer.
 \item Calculate the class scores. If the prediction is correct and the score gap meets the required margin, skip learning for this image. Otherwise, assign positive and negative learning signs to neurons of the correct and highest-scoring incorrect classes.
 \item In every layer, calculate each connection's $A_{ij}$ and $Q_j$ using Equation~\ref{eq:single-image} for a fully connected layer or Equation~\ref{eq:spatial} for a convolutional layer.
 \item Update the weights and histories of neurons with nonzero outputs and learning signs, using Equations~\ref{eq:direction} and either~\ref{eq:update} or~\ref{eq:update-explicit}. Then process the next image.
\end{enumerate}
\end{minipage}

All observations come from the same forward calculation, completed before weights change. No updated layer is used to recalculate another layer's observations for that image. Every layer trains throughout the run; no layer is frozen after an initial training stage. The skip rule still applies to each image, and neurons with no contributing observations keep their weights unchanged.

Each weight uses its own incoming activity, its receiving neuron's output, and its own history. Competition compares activities within a group, and the class signs use the label and whole-network scores. Supervision is therefore shared across the network, but no layer receives an error vector propagated from another layer. There are no activation derivatives, backward passes, or separate classifier fits. The update is not claimed to follow the gradient of a global loss.

\subsection{Minibatch training}
\label{sec:minibatching}
The algorithm above can update after each image. To process several images efficiently on a computer, the experiments instead use minibatches. This subsection defines the minibatch form used for the reported results. Averaging observations before applying the sign operation and updating the history generally produces different weight changes from processing the same images one at a time.

All images in a minibatch pass through the network using the same current weights. The learning signs and observations are calculated separately for each image. Let $b$ index the images, and write $a_{bij}$ and $q_{bj}$ for the values of $A_{ij}$ and $Q_j$ obtained from image $b$ using Equation~\ref{eq:single-image} or~\ref{eq:spatial}.

For neuron or filter $j$, let $\mathcal B_j$ contain the images with a nonzero learning sign for its class and at least one nonzero output from that neuron or filter. Average only over these contributing images:
\begin{equation}
 A_{ij}=\frac{1}{|\mathcal B_j|}\sum_{b\in\mathcal B_j}a_{bij},
 \qquad
 Q_j=\frac{1}{|\mathcal B_j|}\sum_{b\in\mathcal B_j}q_{bj}.
 \label{eq:pooled}
\end{equation}
These averages are used in Equations~\ref{eq:direction} and either~\ref{eq:update} or~\ref{eq:update-explicit}, with one weight and history update per minibatch. If $\mathcal B_j$ is empty, the neuron keeps its weights and history unchanged. Each contributing image has equal weight, even when a convolutional filter has nonzero outputs at different numbers of positions in different images. Thus convolutional observations are averaged first over positions within each image, then over images. With a minibatch of one image, this procedure reduces to the single-image rule above.

\section{Experiments}
\label{sec:experiments}
Both experiments use the same learning rule for MNIST and CIFAR-10, with the settings in Table~\ref{tab:settings}. Both networks train jointly for 100 epochs in minibatches of 32, with each training image presented once per shuffled epoch. The initial step size is $0.00075$, held constant for 20 epochs and then reduced with a cosine schedule over the remaining 80. Neither experiment uses image augmentation or weight averaging.

\begin{table}[!ht]
\centering
\caption{Selected MLP and convolutional network. Every neuron has an assigned class; each layer divides its neurons or filters equally among the ten classes.}
\label{tab:settings}
\begin{tabular}{lll}
\toprule
Setting & MNIST & CIFAR-10 \\
\midrule
Learned layers & 3 fully connected & 3 convolutional \\
Total neurons / channels per layer & $2560,2560,2560$ & $1280,1280,1280$ \\
Neurons / channels per group & $256,256,256$ & $128,128,128$ \\
Outputs kept per group & $128,128,128$ & $8,8,8$ per location \\
Kernel sizes & --- & $3\times3,3\times3,5\times5$ \\
Learned weights, including biases & 15,121,920 & 55,744,000 \\
Required margin $\delta$ & 0.10 & 0.05 \\
Development training / validation & 50,000 / 10,000 & 40,000 / 10,000 \\
Final training / test & 60,000 / 10,000 & 50,000 / 10,000 \\
\bottomrule
\end{tabular}
\end{table}

\paragraph{Inputs and architecture.}
MNIST pixels are scaled to $[0,1]$, centered using the training mean at each pixel, and divided by each image's largest absolute centered value, leaving an all-zero image unchanged. CIFAR-10 pixels are scaled to $[0,1]$ and centered using the training RGB means.

\paragraph{The constant input and learned bias.}
A first-layer $3\times3$ RGB patch contains 27 pixel values. Write these centered values as $p_1,\ldots,p_{27}$. Preprocessing flattens the patch into a vector, appends one extra entry equal to 1, and divides the whole vector by its length:
\begin{equation}
 u=\frac{[p_1,\ldots,p_{27},1]^\top}{\sqrt{\sum_{i=1}^{27}p_i^2+1}}.
 \label{eq:patch-bias-input}
\end{equation}
Each filter has 27 pixel weights and one additional learned weight, its bias $b_j^k$. Its response before competition is
\begin{equation}
 z_j^k=\frac{\sum_{i=1}^{27}w_{ij}^k p_i+b_j^k}{\sqrt{\sum_{i=1}^{27}p_i^2+1}}.
 \label{eq:patch-bias-response}
\end{equation}
The appended 1 is not a pixel or padding. Its weight lets the filter add an offset to its weighted pixel sum; that offset can change which filters win the competition. The same filter weights and bias are used at every patch location. The bias follows the same local learning rule as the other weights, using the final entry of $u$ as its incoming activity. In this first layer that entry is $1/\sqrt{\sum_i p_i^2+1}$ because the entire vector is normalized.

The MNIST first layer likewise appends 1 to its 784 image values and normalizes the resulting 785-entry vector. Deeper layers in both networks append 1 to their incoming activities without this length normalization, so their responses are simply the weighted activity sum plus the learned bias. Weights on nonconstant inputs start as independent Gaussian values normalized to unit length for each receiving neuron; bias weights and running averages start at zero.

The CIFAR-10 network uses convolutions without padding, moving each filter one pixel at a time. Each filter evaluates its ordinary and horizontally reflected orientations, retaining the response with larger absolute value and preserving its sign. The two orientations share weights. Learning maps the selected orientation's observations to those same weights. After each of the first two layers, pooling selects the signed largest-magnitude value in each $2\times2$ window at stride one, then smooths and downsamples using the fixed filter
\begin{equation}
 \frac{1}{16}\begin{bmatrix}1&2&1\\2&4&2\\1&2&1\end{bmatrix}
 \label{eq:smoothing}
\end{equation}
with stride two and reflected boundaries, following the smoothing-before-downsampling approach of \citet{zhang2019}. Reflection does not repeat the boundary value: for example, $[a,b,c]$ is extended to $[b,a,b,c,b]$. The three convolutional outputs have spatial sizes $30\times30$, $13\times13$, and $2\times2$.

\paragraph{Development and final testing.}
Architecture, competition, margin and step size follow the earlier validation scans on the splits in Table~\ref{tab:settings}. The comparison omits dropout and varies only the weight update. Training on each dataset comprised four runs with semi-implicit Lion and four with explicit Lion. The two models in each pair share the same initialization and example order. The four seeds are the original seed selected during development and three additional seeds. Each model trains from scratch on the full training set, including former validation images, with preprocessing statistics computed from training images only. The reported results include every run at epoch 100, evaluated on the same 10,000 official test images for each dataset. These test sets were also used in earlier experiments; the comparison is not a fresh test of the decision to omit dropout.

\subsection{Computational performance}
The training kernels were implemented in Mojo 1.2.0\footnote{\url{https://mojolang.org/}} to allow direct optimization of GPU operations without requiring backpropagation or automatic differentiation. On an NVIDIA GeForce RTX 4090, the original reference runs on the complete training sets averaged approximately 2 seconds per epoch for MNIST and 28 seconds per epoch for CIFAR-10, excluding evaluation and checkpoint saving.

\section{Results and discussion}
Switching from the semi-implicit to the explicit Lion update changed mean test accuracy by $+0.05$ percentage points on MNIST and $-0.02$ on CIFAR-10 (Table~\ref{tab:results}). The 95\% paired $t$ intervals across the four seed differences are $[-0.07,0.17]$ and $[-0.54,0.50]$ points, respectively. They describe variation across seeds on these fixed tests. Neither comparison establishes an advantage for either update, or equivalence between them.

\begin{table}[ht]
\centering
\caption{Official test accuracy (\%) after 100 epochs. Each row summarizes four matched seeds. SD is the sample standard deviation across runs; the endpoints are the lowest and highest observed accuracies.}
\label{tab:results}
\begin{tabular}{llrrr}
\toprule
Dataset & Lion update & Mean $\pm$ SD & Minimum & Maximum \\
\midrule
MNIST & Semi-implicit & $98.14 \pm 0.04$ & 98.10 & 98.19 \\
MNIST & Explicit & $98.19 \pm 0.09$ & 98.06 & 98.27 \\
CIFAR-10 & Semi-implicit & $80.69 \pm 0.28$ & 80.41 & 81.08 \\
CIFAR-10 & Explicit & $80.67 \pm 0.16$ & 80.45 & 80.82 \\
\bottomrule
\end{tabular}

\end{table}

Table~\ref{tab:layer-scores} compares the ordinary sum of three layer scores with each layer scored alone for the semi-implicit models. The sum improves on the final layer alone by an average of 0.06 percentage points on MNIST and 8.68 on CIFAR-10. Earlier layers still compute the features used by deeper layers; only the activities counted in the score change.

\begin{table}[ht]
\centering
\caption{Test accuracy (\%) using different layers for the class score. Entries are means over the four semi-implicit runs per dataset, without retraining.}
\label{tab:layer-scores}
\begin{tabular}{lrr}
\toprule
Activities included in score & MNIST & CIFAR-10 \\
\midrule
Layer 1 alone & 56.93 & 11.19 \\
Layer 2 alone & 91.61 & 32.56 \\
Layer 3 alone & 98.08 & 72.01 \\
Layers 1+2+3 & 98.14 & 80.69 \\
\bottomrule
\end{tabular}

\end{table}

For published context, SCFF reports test accuracies of $98.70\%\pm0.01\%$ on MNIST and $80.75\%\pm0.12\%$ on CIFAR-10 \citep{chen2025scff}. Its features are learned without labels and classified by a separate linear model trained with backpropagation. Channel-wise competitive Forward-Forward reports test errors of $0.58\%\pm0.08\%$ on MNIST and $21.89\%\pm0.44\%$ on CIFAR-10 for its configuration with a learned softmax classifier \citep{papachristodoulou2024}. The Supervised Hebbian Class Network's CIFAR-10 result is close to SCFF's reported accuracy; both published methods report stronger MNIST results than the Supervised Hebbian Class Network. Architectures, preprocessing, supervision, classifiers, training schedules, and computational budgets differ, so these are published reference points rather than controlled comparisons.

The main result is that MLPs and convolutional networks can learn jointly from local activity products when each neuron is assigned a class. That assignment determines which receiving neurons are reinforced or suppressed, while the activities at each connection determine its update. The usual connections let neurons use features computed by neurons assigned to other classes. No separate classifier is needed to translate these representations into a prediction.

\section{Conclusion}
Supervised Hebbian Class Networks provide a method for jointly training MLPs and convolutional networks with a class assigned to every neuron in each learned layer. The relative score gap controls whether learning occurs, and the correct and strongest incorrect classes determine positive and negative Hebbian observations throughout the network. Lion adapted to Hebbian learning combines sign-based updates and local running averages of these observations with an activity-dependent term that controls weight growth. Both Lion updates achieve about 98.2\% mean test accuracy on MNIST and 80.7\% on CIFAR-10. The contribution is a practical form of joint multilayer learning that requires neither gradient calculations nor backpropagation.

\label{main-end}
\bibliography{references}
\bibliographystyle{tmlr}
\end{document}